\documentclass{article}
\usepackage{amssymb}
\usepackage{spconf,amsmath,graphicx}
\usepackage{subfig}
\usepackage{float}
\usepackage{wrapfig}
\usepackage[table]{xcolor}
\usepackage{color}
\usepackage{tikz}
\usepackage{multirow}

\title{Robust and Fully Automated Segmentation of Mandible from CT Scans}
\name{Neslisah Torosdagli$^{\star}$, Denise K. Liberton$^{\dagger}$, Payal Verma$^{\dagger}$, Murat Sincan$^{\dagger}$}
\secondlinename{Janice Lee$^{\dagger}$,  Sumanta Pattanaik$^{\star}$, Ulas Bagci$^{\star}$}
 \address{$^{\star}$ Department of Computer Science, University of Central Florida, Orlando, FL.\\$^{\dagger}$ National Institute of Dental and Craniofacial Research (NIDCR), \\National Institutes of Health (NIH), Bethesda, MD.} 
\begin{document}
%
\maketitle
\begin{abstract}
Mandible bone segmentation from computed tomography (CT) scans is challenging due to mandible's  structural irregularities, complex shape patterns, and lack of contrast in joints. Furthermore, connections of teeth to mandible and mandible to remaining parts of the skull make it extremely difficult to identify mandible boundary automatically. This study addresses these challenges by proposing a novel framework where we define the segmentation as two complementary tasks: recognition and delineation. For recognition, we use random forest regression to localize mandible in 3D. For delineation, we propose to use 3D gradient-based fuzzy connectedness (FC) image segmentation algorithm, operating on the recognized mandible sub-volume. Despite heavy CT artifacts and dental fillings, consisting half of the CT image data in our experiments, we have achieved  highly accurate detection and delineation results. Specifically,  detection accuracy more than $96\%$ (measured by union of intersection (UoI)), the delineation accuracy of $91\%$ (measured by dice similarity coefficient), and less than $1$ mm in shape mismatch (Hausdorff Distance) were found. 
\end{abstract}

\section{Introduction}
\label{sec:intro}
Majority of the head-neck cancers evolve in the craniomaxillofacial (CMF) region. Apart from cancer, more than 16 million Americans need surgical or orthodontic treatment to correct CMF deformities~\cite{zhang2015automatic}. Among all CMF deformities, jaw deformity is the most frequent one. Diagnosis and treatment/surgery planning require precise delineation of CMF bones accurately, and preferably in most efficient ways. The mandible is the lower-jaw bone, the largest, strongest, and the most complex bone serving for the reception of the lower teeth, and channels with blood vessels and nerves. It is important that CMF bones, particularly mandible, to be delineated precisely for better clinical interpretation and treatment guidance~\cite{book}.
However, this procedure is not trivial due to large morphological variations among different patients, disease burden, and inevitable image artifacts including dental fillings, orthodontic wires, bands, and braces. Towards a solution to this challenging problem, there have been many attempts in the literature; however, most of the studies that have addressed the mandible segmentation utilize statistical shape models~\cite{shape_models_papers}. More recently, machine learning based approaches have been proposed~\cite{zhang2015automatic} to avoid segmentation step and focus on measuring clinically useful metrics. In such studies, automatic landmarking process for CMF bones is usually aimed for precise localization of anatomically distinct points. The overall goal in such approaches is to help clinician to make certain geometric measurements easier, faster, while maintaining accuracy. Arguably, we believe that segmentation is still necessary to fully appreciate anatomical information and improve landmark interpretations in high-dimensional space.

\begin{figure}[H]
\centering
\includegraphics[width=\linewidth]{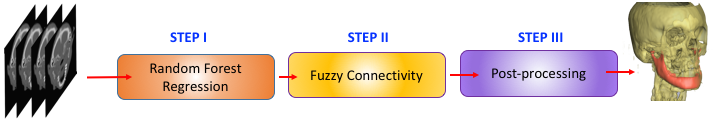}
\caption{Flowchart of the proposed approach is shown. Random Forest regression algorithm is used for mandible recognition (Step I). Gradient-based FC algorithm is then used for delineation of the mandible (Step II). Finally, leak tracing is conducted for improving delineation (Step III).}
\label{fig:pipeline}
\end{figure}

Historically, active contour, shape, and appearance models have been used extensively to model CMF bone shapes, and to extract mandible~\cite{book}. In later years, energy based segmentation methods such as Markov Random Field (MRF) and Conditional Random Fields (CRF) have replaced such methods. In parallel, registration (atlas-based) methods have been reported to achieve relatively higher accuracy when shape and appearance information are integrated. Despite all advances in computational analysis of CMF bones, the problem of accurate and efficient segmentation of mandible with high robustness to CT artifacts remains a challenging task. The main reason is due to the following drawbacks of the current strategies in the literature: 
\begin{itemize}
\item lack of robustness due to existence of varying artifacts in CT images and high anatomical and pathological variations;
\item lack of model-free approaches that do not produce extensive false positives.
\end{itemize}
In this study, we address these two problems by proposing a \underline{data-driven} recognition and delineation approach based on RF Regression analysis and gradient-based Fuzzy Connectivity (FC) image segmentation. To refine the segmented results, we develop a \underline{ state-machine heuristics} approach. 
\section{Method}
We consider the segmentation procedure as consisting of two complementary tasks: recognition and delineation. While recognition is the process of determining roughly ``where" the object is and distinguishing it from other object-like entities in the image, delineation is the act of defining the spatial extent of the object region in the image~\cite{bagci_hsmor}. In our previous publication, we have shown the significance of having a successful recognition step for efficient and accurate segmentation process~\cite{bagci_hsmor, xu2013spatially}.  In this study, an efficient recognition step avoids potential leakages in the delineation by constraining the segmentation region into a tight search space. 

\textbf{Our contributions:} The proposed framework is illustrated in Figure~\ref{fig:pipeline}. Our approach is completely data-driven and no user interaction is required (fully automated). In order to avoid potential leakages, we utilize two methods: first, we constrain the delineation region of mandible into a bounding-box that tightly encloses the mandible with the recognition procedure. We use a RF regression algorithm in a pair programming approach, which helps recognition step to combine information in sagittal, axial, and coronal slices of CT scans in a multi-view settings. Second, we use gradient-based FC algorithm, which has less susceptibility to noise, artifacts, and weak boundaries compared to absolute FC. In the final stage, to handle the potential leakages at the teeth and at the mandible-skull joints, which is impossible to track using any region-based segmentation algorithm, we apply a knowledge-based refinement heuristic strategy, called \underline{a state-machine approach}. In the testing phase, we use MICCAI 2015 Head-Neck segmentation challenge data set, which is clinically classified into 3 groups based on the severity of the CT artifacts. Similar accuracies are obtained in each group regardless of the amount of artifact indicating the robustness of the proposed approach. 
    
\subsection*{Step I: Recognition of mandibular bone}
We assign a probability score for each slice (in multi-view setting) if it includes mandible bone. Then, we combine these views (sagittal, axial, and coronal) to improve probabilistic determination of mandible location. Labeled slices (0 or 1) are used to train an RF-based regression algorithm. Probability scores from each view are fused, and we  obtain a continuous output scale for which probability value of 0.5 and higher for a particular slice is considered to include mandible bone. Instead of multi-view setting, one may directly use 3D implementation of RF based classification. However, particularly in CMF analysis with low dose CT scans, or CT scans with asymmetric pixel sizes,  it is reasonable to control limited information from certain slices and views. 

\subsection*{Step II: Gradient-based fuzzy connectivity}
After identifying the object of interest through a bounding-box, tightly enclosing mandible, we propose to use the gradient-based FC algorithm to minimize leakage possibilities while delineating mandible. FC has shown to be very robust compared to other algorithms such as Graph-Cut (GC) and Level Set (LS)~\cite{jay_math}. However, FC algorithm accepts a pre-defined mean and standard deviation of object's intensity distributions, and since there are bones, joints, and other artifacts in the vicinity of mandible sharing strong intensity similarities, it is still possible for the delineation algorithm to leak. Thus, we re-parametrize the FC algorithm to accept only one parameter that is calculated through gradient image instead of the original gray-scale CT scan. By this change, we emphasize the boundary locations in delineation.

To make the paper self-contained, we briefly explain the core of the FC algorithm in the following. Let a topology on an image is given in terms of an \textit{adjacency} relation ($\mu_\alpha$) such that if $p$ and $q$ are $\alpha$-adjacent to each other, then $\mu_\alpha(p,q)=1$, `0' otherwise. In practice, we set $\alpha=26$ for 3-D mandible bone analysis. While affinity is intended to be a local relation, a global fuzzy relation, called fuzzy connectedness, is induced on the image domain by the affinity functions as described in detail in~\cite{UDUPA1996246}. This is done by considering all possible paths between any two voxels $p$ and $q$ in the image domain, and assigning a strength of FC for each path. The level of the FC between any two voxels $p$ and $q$ is considered to be the maximum of the strengths of all paths between $p$ and $q$. An \textit{affinity relation} $\kappa$ is the most fundamental measure of local hanging togetherness of nearby voxels. For a path $\pi$, which is a sequence of voxels $\langle p_1,p_2,...,p_l\rangle$ with every two successive voxels being adjacent, given \textit{fuzzy affinity function} $\mu_\kappa(p_i, p_{i+1})$, the strength of the path is defined as the minimum affinity along the path~\cite{xu2013spatially, UDUPA1996246}:
\begin{equation}
\mu_{\mathcal{N}}(\pi)=\min_{1\leq i<l}{\mu_\kappa(p_i,p_{i+1})}.
\end{equation}
Then, the strength of connectedness $\mu_\mathcal{K}(p,q)$ between any two voxels $p$ and $q$ is the strength of the strongest path between them as
\begin{equation}
\mu_\mathcal{K}(p,q)=\max_{\pi\in\mathcal{P}(p,q)}{\mu_\mathcal{N}(\pi)},
\end{equation}
where $\mathcal{P}(p,q)$ denotes the set of all paths between $p$ and $q$. A FC object $\mathcal{O}$ (i.e., mandible) in an image can be defined for a predetermined set of seeds $S$. Note that in our algorithm, there is only one seed necessary to initiate FC delineation and we solve this issue by selecting the voxel with highest intensity value and largest connected component in the bounding-box. 
The final object is obtained by thresholding over the fuzzy object $\mathcal{O}$ for strength of connectedness.

\begin{figure*}
\includegraphics[width=1\linewidth]{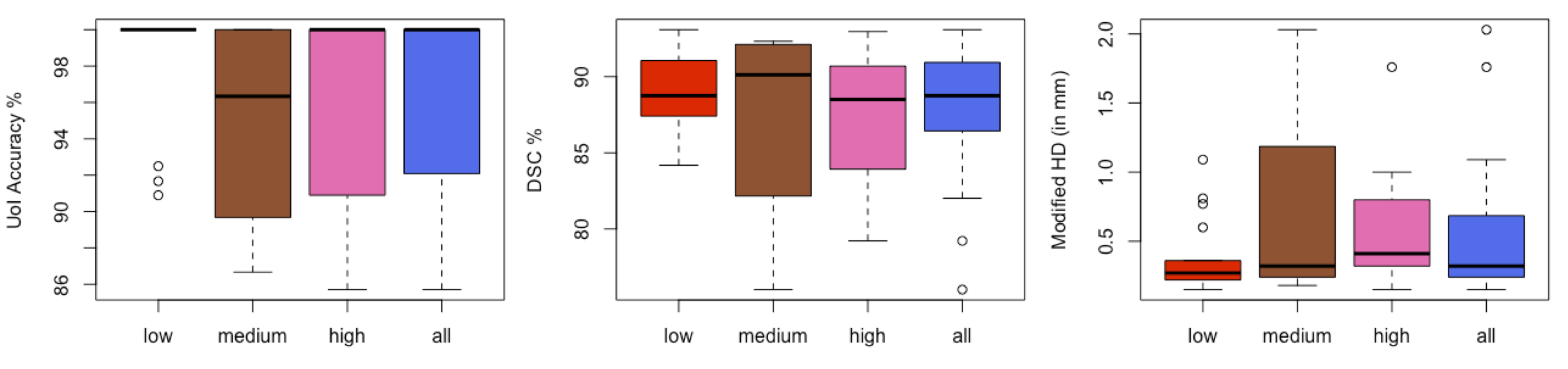}
\caption{UoI, DSC, and modified Hausdorff Distance (HD) metrics are shown for evaluation of detection and segmentation methods with respect to varying amount of artifact (x-axis: low, medium, high, and overall). Note that, high UoI indicates better recognition, lower HD and higher DSC scores indicate accurate segmentation.}
\label{fig:boxplot}
\end{figure*}

\begin{figure}[H]
\centering
\includegraphics[width=\linewidth]{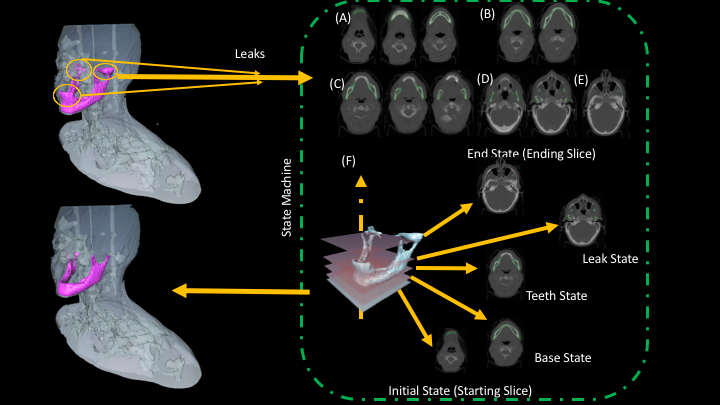}
\caption{Boundary refinement using a five step state-machine.}
\label{fig:stateMachine}
\end{figure}

\begin{figure}[H]
\centering
\includegraphics[width=1\linewidth]{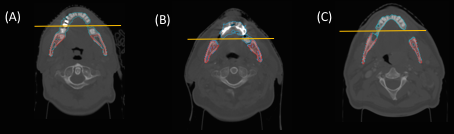}
\caption{Separation of teeth and mandible, ground-truth (red) and FC (blue).}
\label{fig:teeth}
\end{figure}

\subsection*{Step III: Boundary refinement for final segmentation}
\begin{figure}[H]
\centering
\includegraphics[width=1\linewidth]{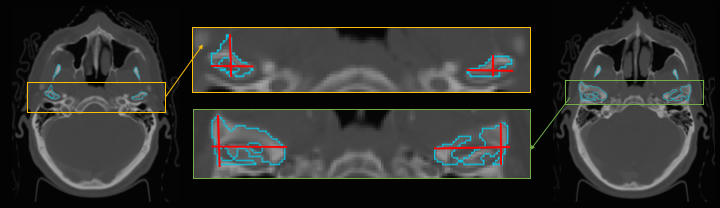}
\caption{Leak state, FC (blue) and the width and height of delineated mandible (red). Abrupt change in width or height in consecutive slices means a leak in the mandible skull joint.}
\label{fig:leak}
\end{figure}

We present a new boundary refinement algorithm to handle leaks. Although it is small, leakage can still occur during this separation due to strong overlap between intensity distribution of these structures (Figure ~\ref{fig:stateMachine}). Based on anatomical knowledge, we design a state-machine, composed of five states: initial state, base state, teeth state, leak state, and ending state (Figure~\ref{fig:stateMachine}). The number of connected components and their sizes in consecutive axial slices are used to switch from one state to another. Briefly, we utilize a state-machine heuristic approach to track potential leak, and then refine those regions by incorporating the domain knowledge: axially, mandible regions include a few connected components with small sizes and this information is quite robust even with highly deformed jaws (initial state in Figure~\ref{fig:stateMachine}-A). Following, the connected components start merging and becomes large in physical size (base state in Figure~\ref{fig:stateMachine}-B). Next, the number of connected components increases as teeth are introduced (teeth state in Figure~\ref{fig:stateMachine}-C). In the teeth state, \textit{k-means} clustering with two classes is used to separate teeth from mandible. This rough separation allows us to automatically localize teeth, located in the front portion of the axial slice, and mandible at the back (See Figure~\ref{fig:teeth}). In the leak state,  we use the change in width and height of delineated mandible to track \textit{abrupt changes} to detect leak areas (leak state in Figure~\ref{fig:stateMachine}-E  and Figure~\ref{fig:leak}).

\section{Experiments and Results}
\textbf{Data:} We clustered MICCAI Head-Neck Challenge 2015 data set into 3 groups, based on the severity of artifacts on the CT data as: No or Low, Medium, and High. This was done to experimentally show robustness of our proposed system with respect to the varying amount of artifact. While 17 volumes were classified as no or low artifact, 4 of them were labeled as medium, and 19 of them were clinically classified as high amount of artifacts. The in-plane resolution of the CT scans was 1.12 x 1.12 mm, and the slice thickness was 3 mm. 

We conducted our experiments on a low-end regular laptop, MacOS-X (2.6 GHz Intel Core-i5) with 8GB CPU memory. Three CT volumes were randomly selected for training and the rest were used for testing in RF  based recognition experiments. Based on commonly used detection/recognition evaluation metrics, called union of intersect (UoI), we have performed a detection accuracy of more than $96\%$. For delineation accuracy, we have used dice similarity coefficient (DSC) as well as a modified version of Hausdorff Distance (HD) to quantify boundary shape mismatch. The delineation performances are summarized for all CT scans and with respect to the varying amount of CT artifacts (low, medium, and high) in Figure~\ref{fig:boxplot}.  

Compared to the existing work of ~\cite{challange_winner}, the winning algorithm of the MICCAI 2015 Head-Neck segmentation challenge, our algorithm provided similar DSC and HD metrics for all CT scans regardless of the amount of artifacts, confirming the robustness of the proposed method. Furthermore, we claim at least two superiorities: \textbf{(1)} our algorithm is data-driven, \textbf{(2)} regardless of the amount of artifacts, we obtained similar accuracies for mandible segmentation, hence the proposed algorithm is robust.

\section{Discussions and Concluding Remarks}
In this work, we develop a data-driven, robust, and accurate method for automatic mandible detection and delineation from CT scans. The proposed approach is based on RF regression algorithm for determining rough location of the mandible, and region-based segmentation algorithm with automatic initialization. We integrate anatomy knowledge in the final stage as a state-machine procedure. With this refinement, fine-tuning of the delineation is performed successfully regardless of the amount of CT artifacts and high anatomical variability in the mandible bones. This verifies the robustness of the overall system. By using publicly available segmentation challenge data (MICCAI 2015- Head-Neck Data Set), we tested and evaluated the performance of the recognition and delineation steps of the proposed framework, and obtained similar accuracies compared to the work of ~\cite{challange_winner} but with superiority in robustness, efficiency, and fully-automatic nature.

Although the proposed algorithm is promising in several ways, following limitations of our work should be noted as well. Performance of our algorithm is not tested in CT scans with bone fractures or bone-loss in or nearby the mandible. Although these cases may be rare in this particular settings, it may be desirable for clinicians to handle such cases (challenge data-set is pertaining head-neck cancer) with the software tool as well.  Furthermore, in our study, we integrated expert knowledge on human anatomy (specifically mandible) in state-machine model for the refinement step. However, for congenital diseases and trauma cases, modeling expert knowledge may not be accurate for leak tracking. Therefore, our future studies will explore potential ways of detecting pathological changes in the bones as a part of whole segmentation procedures. 

\bibliographystyle{IEEEbib}
\bibliography{refs}
\end{document}